\documentclass[11pt,a4paper]{article}
\usepackage[hyperref]{acl2019}
\usepackage{times}
\usepackage{latexsym}
\usepackage{arydshln}
\usepackage{url}
\usepackage{lineno}
\usepackage{times}
\usepackage{color}
\usepackage{bm}
\usepackage{latexsym}
\usepackage{amssymb,amsmath}
\usepackage{mathrsfs}
\usepackage{graphicx}
\usepackage{caption}
\usepackage{subfigure}
\usepackage{url}
\usepackage{caption}
\usepackage{tabularx}
\aclfinalcopy % Uncomment this line for the final submission
\def\aclpaperid{2376} % Enter the acl Paper ID here
\usepackage{booktabs}
\usepackage{mathrsfs}
\usepackage{makecell}
\usepackage{multirow}
\usepackage{todonotes}
\usepackage{enumitem}
\usepackage{amsmath}
\usepackage{algorithm}
\usepackage[noend]{algpseudocode}

\algrenewcommand\algorithmicforall{\textbf{foreach}}
\algrenewcommand\algorithmicindent{.8em}

\usepackage[normalem]{ulem}

\newcommand{\bv}{\boldsymbol{v}}
\newcommand{\bu}{\boldsymbol{u}}

\newcommand{\bx}{\boldsymbol{x}}

\newcommand{\bp}{\boldsymbol{p}}

\def\mW{{\bm{W}}}
\def\mX{{\bm{X}}}

\title{Towards Scalable and Reliable Capsule Networks\\for Challenging NLP Applications}

\author{Wei Zhao$^{\dagger}$, Haiyun Peng$^{\ddagger}$, Steffen Eger$^{\dagger}$, Erik Cambria$^{\ddagger}$ and Min Yang$^{\Phi}$\\
{$^\dagger$ Computer Science Department, Technische Universit\"at Darmstadt, Germany}\\
{$^{\ddagger}$ School of Computer Science and
Engineering, Nanyang Technological University, Singapore}\\
{$^{\Phi}$ Shenzhen Institutes of Advanced Technology, Chinese Academy of Sciences, China}\\
{\tt www.aiphes.tu-darmstadt.de}
}
\date{}

\begin{document}
\maketitle
\begin{abstract}
Obstacles hindering the development of capsule networks for challenging NLP applications include poor scalability to large output spaces and less reliable routing processes.
In this paper, we introduce (\romannumeral1) an agreement score to evaluate the performance of routing processes at instance level; (\romannumeral2) an adaptive optimizer to enhance the reliability of routing; (\romannumeral3) capsule compression and partial routing to improve the scalability of capsule networks. We validate our approach on two NLP tasks, namely: multi-label text classification and question answering. Experimental results show that our approach considerably improves over strong competitors on both tasks. In addition, we gain the best results in low-resource settings with few training instances.\footnote{Our code is publicly available at \href{http://bit.ly/311Dcod}{http://bit.ly/311Dcod}} 
\end{abstract}

\section{Introduction}\label{sec:introduction}
In recent years, deep neural networks have achieved outstanding success in natural language processing (NLP), computer vision and speech recognition. However, these deep models are data-hungry and generalize poorly from small datasets, very much unlike humans~\cite{Lake2015}. 

This is 
an important issue in NLP 
since 
sentences with different surface forms can convey the same meaning (paraphrases) 
and 
% not all of which 
% with meaning-variations in paraphrases 
% can be enumerated 
not all of 
them 
%paraphrases 
can be enumerated
% such paraphrases with meaning-variations are looming shortage 
in the training set. For example, 
% \textit{Peter was responsible for the library} and \textit{Peter was in charge of the library} 
 \textit{Peter did not accept the offer} and \textit{Peter turned down the offer}
are semantically equivalent, but %they are 
%written in different surface forms.
use different surface realizations. 

\iffalse
\begin{figure}
\centering
\includegraphics[width=.6\linewidth]{figures/capsule-crop.pdf}
\caption{The capsule $T_i$ holds one vector representing a bigram \textit{Peter was}, and its probability $p_i$ shows if it is useful for the task at hand.}
% for one bigram "Peter was", and its probability $p_i$ shows if it is useful for the task at hand.}
\label{fig:capsules}
\vspace{-0.1in}
\end{figure}
\fi

\begin{figure}
\centering
\includegraphics[width=\linewidth]{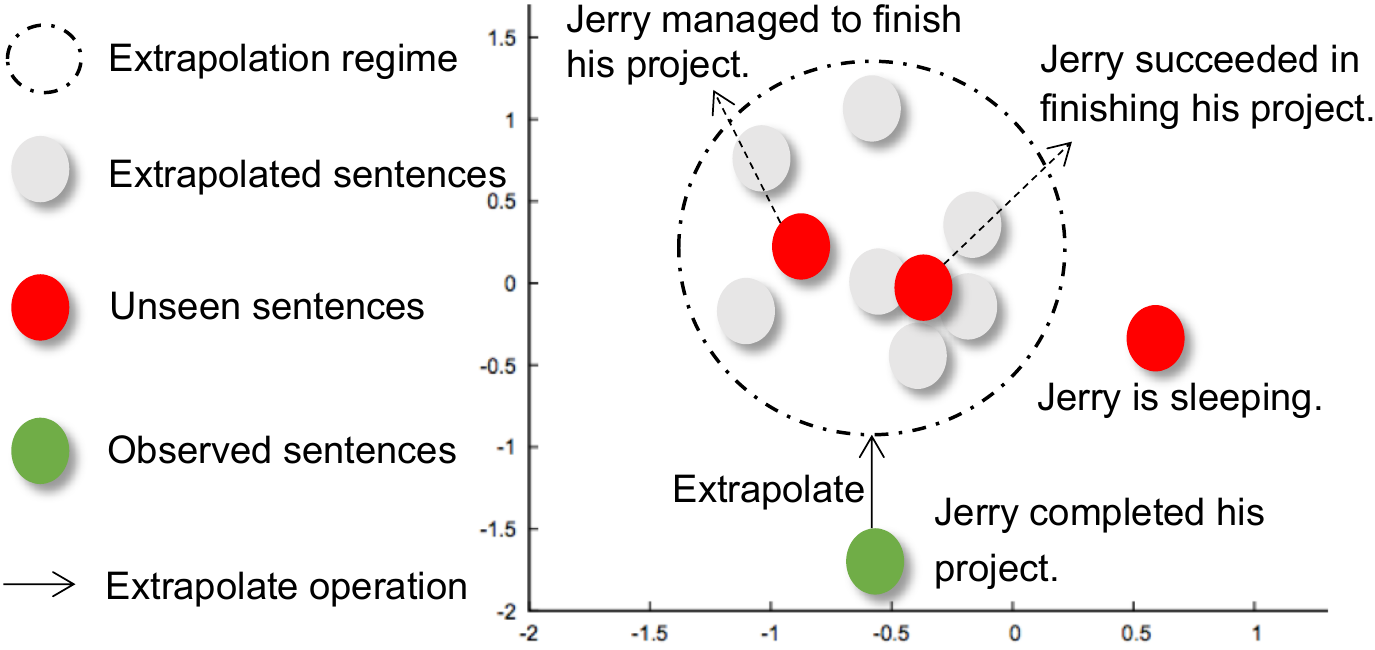}
\caption{
% An illustration of how observed sentences are generalized to unseen sentences that have similar meaning but different surface forms. 
% As a consequence, 
% The unseen sentences could be generalized successfully as long as their embeddings are found in the extrapolation regime.
The extrapolation regime 
for an observed sentence can be found during training. Then, the unseen sentences in this regime may be generalized successfully.
% as long as their embeddings are in this regime.
% Those sentences would be extrapolated from observed sentences and then aggregated by high-dimensional agreement, as the way of clustering.
}
% Generalize unseen sentences by extrapolations on observed ones followed by strong agreement, as the way of clustering, where 'US' and 'OS' are the abbreviation of "unseen sentences" and "observed sentences".}
% The sentences in green color would not pose massive extrapolation to generalize to unseen sentences in gray color, where cross arrows stand out of them to represent those unseen sentences by high dimensional agreement.

% The representations of unseen sentences in gray color can be extrapolated by the sentences that have similar meaning in green color.}
\label{fig:extrapolation}
\vspace{-0.15in}
\end{figure}

%Impressive 
In image classification, progress on the generalization %cap
ability of 
deep networks 
%for image classification 
has been made by 
% has been made on \yangmin{improving the generalization capability of image classification models by using capsule networks} generalization capability in image classification 
% from little data 
% by using 
capsule networks~\cite{sabour2017dynamic,hinton2018matrix}. They %capsule-based 
%models 
are capable of generalizing to the same object in different 3D images with various viewpoints.

%with 
% in different viewpoints like 3D images.
% \todo{SE: I'm not sure what you mean. ``They are viewpoint invariant, i.e., can generalize from one image to the same image in another viewpoint''? Is it important that the images are 3d rather than 2d?}

Such generalization capability can be learned from examples with few viewpoints by extrapolation~\cite{hinton2011transforming}.
% being invariant 
% In terms of generalization capability from little data, 
% capsule networks~\cite{sabour2017dynamic}, 
% % as proposed by \todo{I think you could directly remove ``as proposed by"}~\citet{sabour2017dynamic}, 
% stand out of other deep learning competitors. %in classification tasks. 
% %Capsule networks, as proposed by~\citet{sabour2017dynamic}, %can play a critical
% %have the potential of 
% %role in 
% They can 
% %improving generalization capability 
% improve generalization from little labeled data due to their \textit{viewpoint invariance} property \todo{you could consider to merge this sentence with the last sentence?}. 
%Inspired by their success in computer vision, 
%we hypothesize that capsule networks are capable of generalizing to the texts with semantically similar meaning but different surface forms. 
This %generalization ability 
suggests that capsule networks can similarly abstract away from different surface realizations in NLP applications. 
% meaning-variations of different surface forms in texts. 

Figure~\ref{fig:extrapolation} illustrates this idea of 
how observed sentences in the training set are generalized to unseen sentences by extrapolation.
% To further improve generalization for unseen sentences, capsules employ massive extrapolation followed by high dimensional agreement.\todo{SE: this sounds like complete exaggeration. `massive extrapolation'}\wei{Nope. For one unseen sentence, capnets yield thousands "prediction capsules" for massive extrapolation. I will make this clear}
%, which will be discussed in next section.
In contrast, traditional neural networks require massive amounts of training samples for generalization. This is especially true in the case of convolutional neural networks (CNNs), 
% to generalize to unseen word and sentences, 
where pooling operations
% make such neural models 
wrongly discard positional information and do not consider hierarchical relationships between local features~\citep{sabour2017dynamic}.

% they lose almost all information and their hierarchical relationships by~\citee{sabour2017dynamic}. 

\begin{figure}[b]
\centering
\includegraphics[width=\linewidth]{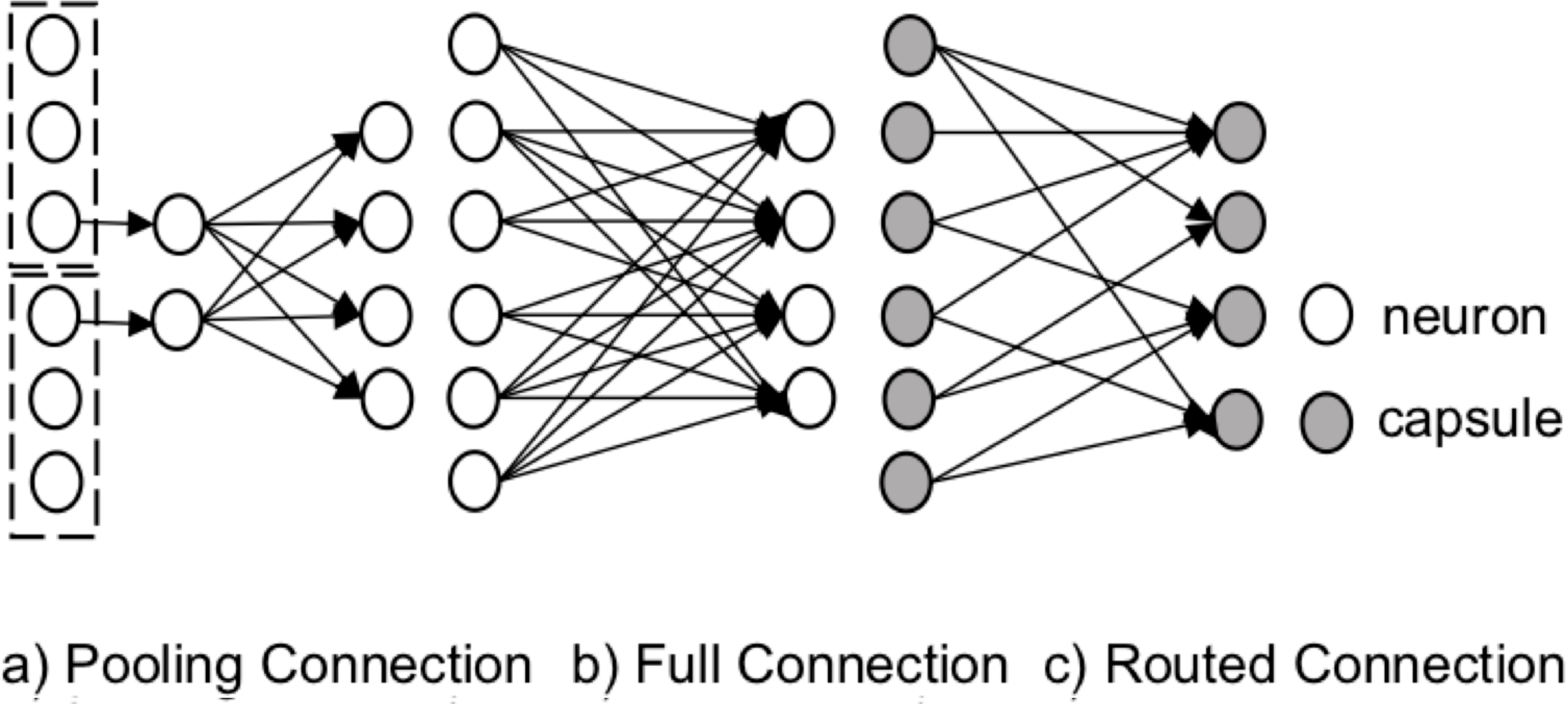}
\caption{Outputs attend to a) active neurons found by pooling operations b) all neurons c) relevant capsules found in routing processes.
% coincidence.
}
\label{fig:layers-comparision}
\vspace{-0.1in}
\end{figure}

Capsule networks, instead, have the potential for learning hierarchical relationships between consecutive layers by using routing processes without parameters, which are clustering-like methods~\cite{sabour2017dynamic} and additionally
% which \todo{SE: additionally?} 
improve the generalization capability. We contrast such routing processes with pooling and fully connected layers %, as shown 
in Figure~\ref{fig:layers-comparision}.

Despite some recent success in NLP tasks~\cite{wang2018towards,xia2018zero,xiao2018mcapsnet,zhang2018attention,zhao2018investigating}, a few %substantial 
important 
obstacles still hinder the development of capsule networks for mature NLP applications.
% (e.g., datasets with large output spaces as in language modeling).

\begin{figure}
\begin{minipage}{0.49\linewidth} 
	\centerline{\includegraphics[width=\linewidth,height=1.25in]{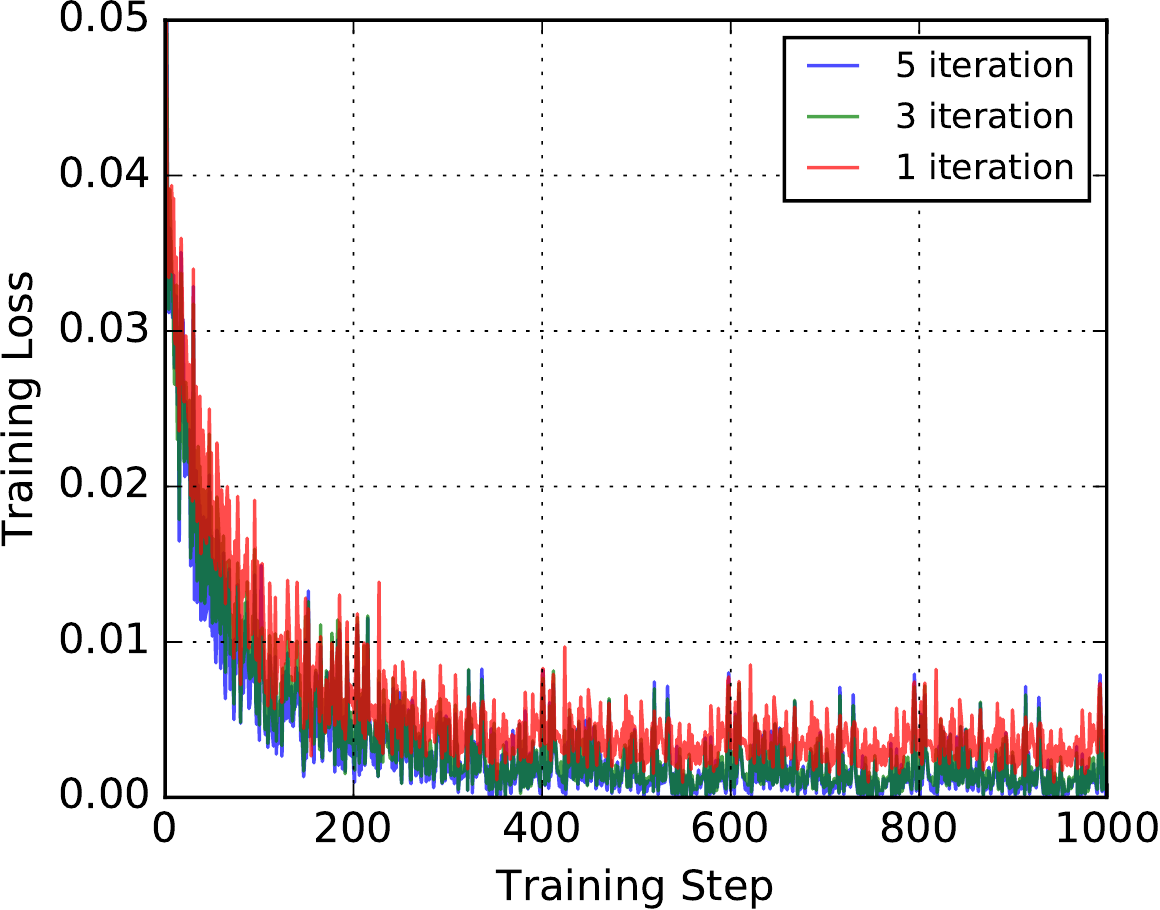}} 

\end{minipage} 
\begin{minipage}{0.49\linewidth} 
	\centerline{\includegraphics[width=\linewidth]{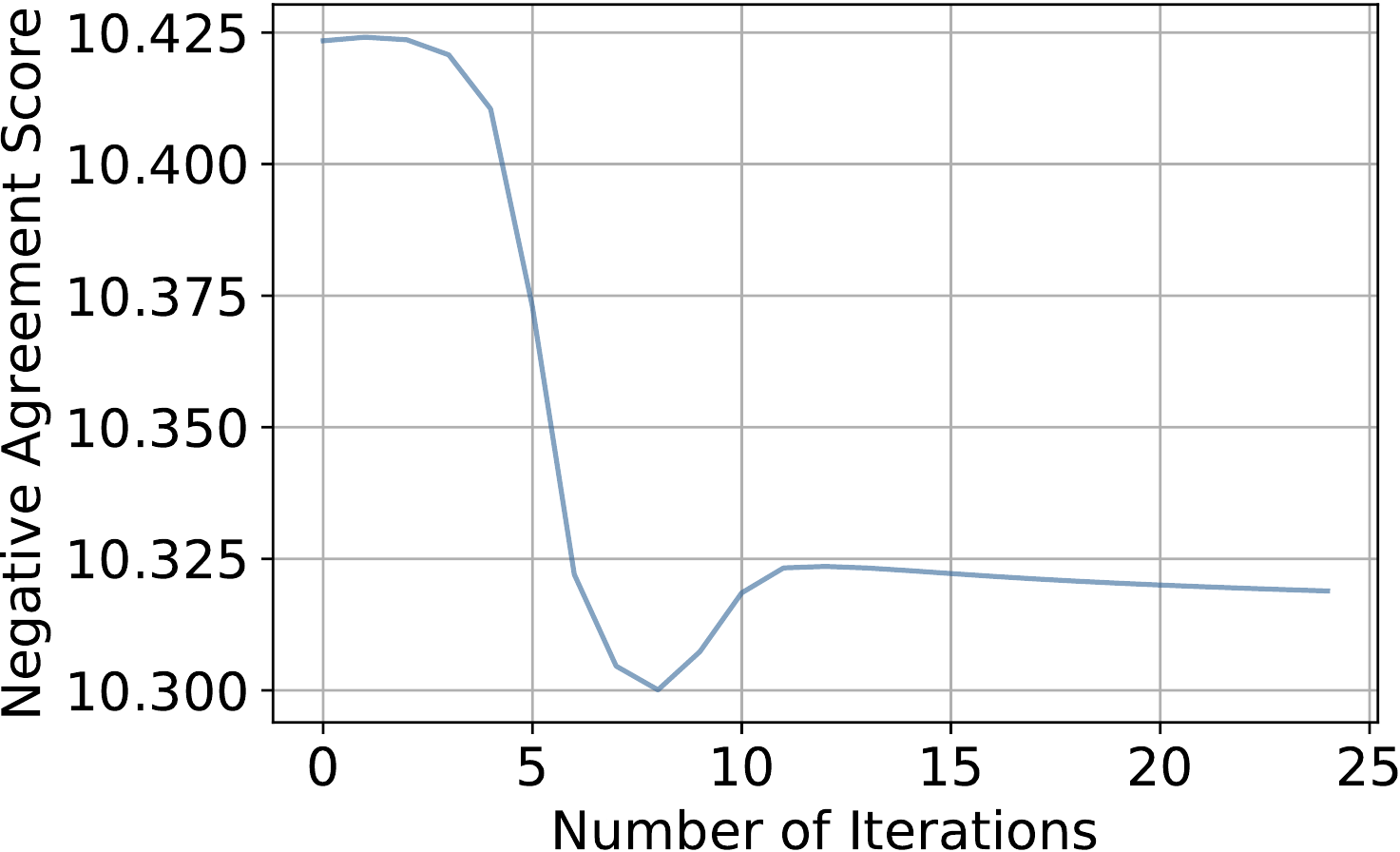}}

\end{minipage} 
 \caption{left) System-level routing evaluation on all examples; right) Instance-level routing evaluation on one example. 
% (The negative agreement score will be discussed later)
 }
\label{fig:evaluate-routing}
\vspace{-0.15in}
\end{figure}

% existing capsule networks are still in their infancy due to the following reasons: 
%(\romannumeral1) 
% existing routing algorithms with a fixed number of iteration can converge to a lower training loss, while 
% which raises the risk of undesirable non-convergence for many examples
% the convergence of routing process is not carefully studied However, we reveal that routing process with a fixed number 
% can converge to a lower training loss, but
% whether routing process get converged by watching training loss
For example, selecting the number of iterations is crucial for routing processes, because they iteratively route low-level capsules to high-level capsules in order to learn hierarchical relationships between layers. However, existing routing algorithms %imposes the risk to instability.
% like dynamic routing~\citet{sabour2017dynamic} and EM routing~\citet{hinton2018matrix}
%using 
use the same number of iterations for all examples, which is not 
% optimal: %reliable:
reliable to judge the convergence of routing.
%\todo{SE: why is this not reliable? Wei: explained in Figure 3, causing individual examples non- converged at instance-level. SE: "reliable" has a certain meaning, but probably not the intended one.} 
As shown in Figure~\ref{fig:evaluate-routing}, a routing process with five iterations on all examples converges to a lower training loss at system level, but 
%if we take a closer look at 
on instance level %convergence on 
for one example, %it still doesn't get converged after five iterations 
convergence has still not obtained. 

Additionally, training capsule networks is more difficult than traditional neural networks like CNN and long short-term memory (LSTM) due to the large number of capsules and potentially large output spaces,
% \todo{SE: LSTMs also may have large output spaces. Maybe routing is the culprit?} 
which 
% output space is as to the tasks, it doesn't matter with models themselves. for example, language model has large output space (vocabulary), news categorization has small output space 
%results in the needs for extensive computational resource in routing processes. 
requires extensive computational resources in the routing process. 
% Such issue prevents capsule networks from mature NLP applications.

In this work, we address these issues via the following contributions:
% Our contributions can be summarized as follows: 
\begin{itemize}[topsep=5pt,itemsep=0pt,leftmargin=*]

\item We formulate routing processes as a proxy problem minimizing a total negative agreement score in order to 
% by providing a clear objective function, 
% and expect it can be used to 
evaluate how routing processes perform 
% the convergence of routing processes 
at instance level, which will be discussed more in depth later.

\item We introduce an adaptive optimizer to self-adjust the number of iterations for each example in order to improve instance-level convergence and enhance the reliability of routing processes.

\item We present capsule compression and partial routing to achieve better scalability of capsule networks on datasets with large output spaces.

% improve the scalability of capsule networks using capsule compression and partial routing for datasets with large output space.

% It is supervised because common neural networks need massive training examples and sophisticated models with more parameters to improve generalization,
% \todo{SE: why even?}\wei{for common neural networks, they need massive training examples and sophisticated networks with more parameters to improve generalization}
\item Our framework outperforms strong baselines on multi-label text classification and question answering. We also demonstrate its superior generalization capability %in fewer training examples condition.
in low-resource settings. % with few training examples. 

% and parameters

% our experimental results show solid performance gains over strong competitors on large multi-label text classification and question answering.
% in several NLP tasks such as large multi-label text classification and question answering. 
% show that our proposed model achieves competitive results over strong competitors on multi-label datasets.

% \item Above all, we %exhibit the 
% show the superior generalization capability of capsule networks with even fewer training examples and parameters.
\end{itemize}

%\todo{SE: this is uncessarily repetitive}
% The rest of the paper is organized as follows.
% \S\ref{sec:approach} introduces our NLP-Capsule framework. 
% \S\ref{sec:setup} and \S\ref{sec:experiments} outline our experiments.
% \S\ref{sec:related} presents related work on MTC and capsule networks.  %Finally, 
% % \S\ref{sec:discussion} discusses the scalability of our capsule network. 
% \S\ref{sec:conclusion} concludes the paper.

\section{NLP-Capsule Framework}\label{sec:approach}

% In this section, we elaborate our model, depicted in Figure~\ref{fig:1}. It is a variant of the capsule networks proposed in~\citet{sabour2017dynamic}. It consists of four layers: a document-level convolutional layer, a primary capsule layer, an aggregation layer, and a class capsule layer.

\begin{figure*}
\centering
\includegraphics[width=\linewidth]{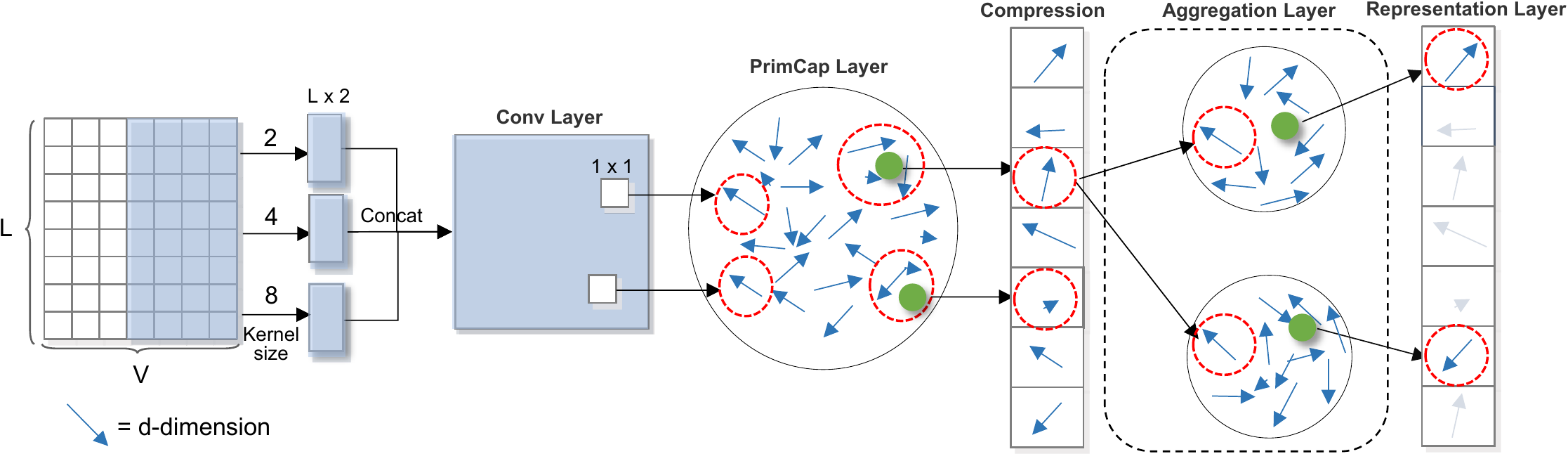}
\caption{An illustration of NLP-Capsule framework.}
\label{fig:1}
\vspace{-0.15in}
\end{figure*}

We have motivated the need for better capsule networks being capable of scaling to large output spaces and higher reliability for routing processes at instance level. We now build a unified capsule framework, which we call NLP-Capsule. It is shown in Figure~\ref{fig:1} and described below.

% In this section, we take inspiration from Capsule Networks~\cite{sabour2017dynamic} in computer vision, and build a unified capsule framework, NLP-Capsule, for two NLP tasks. 
%, namely NLP-Capsule.
%in Figure~\ref{fig:1}, and its applications will be discussed in the next section.
% Its architecture is shown in Figure~\ref{fig:1} and described below. 
%\todo{SE: is it singular or plural? Check each case}

\subsection{Convolutional Layer} 
%Unlike the N-gram convolutional layer in~\citet{zhao2018investigating}, 
We use a convolutional operation 
to extract features from documents by taking a sliding window over document embeddings.

%Let $\mX \in \mathbb R^{l \times v}$ be a set of $v$-dimensional word embeddings as to an input document of size $l$. 
Let $\mX \in \mathbb R^{l \times v}$ be a matrix of stacked $v$-dimensional word embeddings for an input document %of size 
with $l$ tokens. 
% Each element $\bx_{i}$ denotes a word embedding corresponding to the $i$-th word in the current document.
% where $L$ is the length of the document and $V$ is the embedding size of words. 
% We denote $\bx_{i} \in \mathbb R^V$ as a $V$-dimensional word embedding as to the $i$-th word in the current document. %A convolution filter 
Furthermore, let $\mW^a \in \mathbb R^{l\times k}$ be a convolutional filter with a width $k$. We apply this filter
% , where $K$ is the width of the filter 
% (typically $K\sim$10?)
%is %the number of few dimensions of embedding 
%\todo{SE: number of possible dimensions?} 
% the size of the \wei{(2,4,8)}\todo[inline]{SE: say what $K$ is!! ``embedding dimension'' is not appropriate --- that's already called V}
% while sliding over the entire document. The filter $\mW^a$ is 
to a local region $\mX^{\intercal}_{i:i+k-1} \in \mathbb R^{k\times l}$ to generate one feature: 
% (with stride of 2) 
%\todo{SE: what is $T$?} \wei{$T$ is transpose}
%\todo[inline]{SE: unclear to me which mathematical conventions you are using. I assume $i:i+K-1$ is the set $\{i,i+1,\ldots,\}$. In standard math, this is denoted by square brackets. What's the meaning of your $\times$ operator? This $\matbf{x}$ with the subscripts, it's a matrix, right?} 
% with:
\begin{align*}
m_i = f(\mW^a \circ \mX^{\intercal}_{i:i+k-1})  
\end{align*}
where $\circ$ denotes element-wise multiplication, and $f$ is a nonlinear activation function (i.e., ReLU). For ease of exposition, we omit all bias terms. 

Then, we can collect all $m_i$ into one feature map %$\{
$(m_1,\ldots,m_{(v-k+1)/2})$ %\}$ 
after %taking the filter slider 
sliding the filter 
over the current document. To increase the diversity of features extraction, we concatenate multiple feature maps extracted by three filters with different window sizes (2,4,8) and pass them to the primary capsule layer.

\subsection{Primary Capsule Layer}
In this layer, %\todo{SE: a?} 
we use a group-convolution operation to transform feature maps into primary capsules.
As opposed to using a scalar for each element in the feature maps,
% in the convolutional layer, 
capsules use a group of neurons to represent each element in the current layer,
% a group of neurons stands for 
% each element in the primary capsule layer is a group of neurons, 
which has the potential for preserving more information.
% Each primary capsule include a group of neurons instead of scalar to preserve more information.

% transforming input feature maps into a set of primary capsules, expecting to preserve 
% preserve linguistic properties of the input feature map %(e.g., semantics, morphology)

% with a group-convolution operation. 
%\todo{SE: extract?} \todo{SE: again the instantiated parameters :D}. 
% By this way, capsules can preserve more information by extracting the features from a document region into one vector.
Using $1 \times 1$ filters $\mW^b=\{w_1,...,w_d\} \in \mathbb R^d$, in total $d$ groups are used to 
% produce one capsule $\bp_i$, which 
transform each scalar $m_i$ in feature maps to one capsule $\bp_i$, a $d$- dimensional vector, denoted as:
% while sliding over each element $m_i$ in the feature maps (with stride of 1). The capsule $\bp_i$ is defined by
% Sliding over the feature map $\mathbf M=[m_{ij}]$ (with stride of 1), $1 \times 1$ filters $w=\{w_1,...,w_d\} \in \mathbb R^d$ in total $d$ groups produce one capsule $p_i$ by 
% \todo{SE: this is unreadable. To start with: don't use a participle. Make it an active sentence: Filters [...] slide over the feature map [...]}:
% These capsules comprise a channel $P_j$ of primary capsule layer.
\begin{align*}
\bp_i = g(p_{i1}\oplus p_{i2} \oplus \cdots \oplus p_{id}) \in \mathbb R^d
\end{align*}
where $p_{ij}=m_{i} \cdot w_j \in \mathbb R$ and $\oplus$ is the concatenation operator.
%\todo[inline]{SE: $m_i$ is a scalar and so is $W_j^b$?} %\wei{Yes, $W_j^b$ is scalar}, 
% $d$ is the dimension of the capsule 
Furthermore, $g$ is a non-linear function (i.e., squashing function).
%\todo[inline]{SE: what is $\oplus$?}\wei{$\oplus$ is the concatenation operator} 
The length $||\bp_i||$ of each capsule $\bp_i$ 
%\todo[inline]{SE: what is the intensity of capsule?} \wei{the norm of capsule}
%shows 
indicates the probability %showing whether 
of it being useful for the task at hand.
%\todo{SE: class label? Any class label?}. 
Hence, a capsule's length has to be constrained into the unit interval $[0,1]$ by the squashing function $g$: %\todo[inline]{SE: there's no $g$ here} \wei{in Eq. 2}
% \todo[inline]{SE: very messy. What is $s_j$? Is this the definition of $g$?}
\begin{align*}
% \bu_j = \g(\bp_j)=\frac{||\bp_j||^2}{1+||\bp_j||^2}\frac{\bp_j}{||\bp_j||}
g(\bx)=\frac{||\bx||^2}{1+||\bx||^2}\frac{\bx}{||\bx||}
\end{align*}
% where $\bu_j$ is one normalized capsule.

\paragraph{Capsule Compression} 

One major issue in this layer is that the number of primary capsules becomes large %to some extent along with the increased length of 
in proportion to the size of the 
input documents, which %results in the needs for extensive 
requires extensive computational resources
% poses increased computational demand in routing processes 
% costs high 
in routing processes 
(see Section~\ref{sec:agg}). To mitigate this issue, we %propose a compression operation to 
condense the large number of primary capsules into a smaller amount. In this way, we can merge similar capsules and remove outliers.
% since we observed that certain 
%of 
% primary capsules may be noisy or similar to each other.
% \todo{SE: now you give three reasons: (i) they may be noisy, (ii) they may be similar to each other, (iii) to reduce computational costs} 
Each condensed capsule $\bu_{i}$ is calculated by using a weighted sum over all primary capsules, denoted as: 
% We take a simple solution to capsule compression operation by taking a weighted sum over all primary capsules, denoted as: 
% scalability of capsule networks due computation cost of routing processes in the aggregation layer (See section 2.3). Consequetly, we present a capsule compression operation to yield very few capsules,
% by taking a weighted sum over 
% all primary capsules since we discovered that some primary capsules might be noisy or similar to each other, We denoted the compression as:
%\vspace{-0.01in}
\begin{align*}
\hat \bu_{i}=\sum_{j} b_{j} \bp_{j} \in \mathbb R^d  
\end{align*}
where the parameter $b_{j}$ is learned by supervision.
% $\hat \bp_{i}$ is one compressed capsule, 
% $b_j \in \mathbb{R}$ is a parameter learned by supervision.
% \todo{SE: what is the sum over?} 
% The number of high-level capsules is supposed to be far smaller than the %sheer 
% number of primary capsules. 

\subsection{Aggregation Layer}\label{sec:agg}
Pooling is the simplest aggregation function routing condensed capsules into the subsequent layer, 
% which only passes the prominent ones,
but it loses almost all information during aggregation. 
% wrongly discards positional information.
% computational efficient ways to aggregate these condense capsules between two layers, but they do not consider the hierarchical relationships and discard positional information. 
Alternatively, routing processes are introduced to iteratively route condensed capsules into the next layer
% \todo{SE: the above??? the next? the following? the subsequent? Wei: next..} 
for
% capsule networks aggregate these low-level capsules into high-level capsules in the next layer by routing processes, clustering-like methods, 
learning hierarchical relationships between two consecutive layers.
We now describe this iterative routing algorithm. 
% use routing processes, clustering-like methods, to route these low-level capsules into the higher layer in order to learn hierarchical relationships between two consecutive layers.
% route these low-level capsules into the higher layer by using routing processes, clustering-like methods. 
Let $\left\{\bu_{1}, \ldots, \hat \bu_{m}\right\}$ and $\left\{ \bv_{1}, \ldots, \bv_{n}\right\}$ be a set of condensed capsules in layer $\ell$ and a set of high-level capsules in layer $\ell+1$, respectively. The basic idea of routing is two-fold.

First, we transform the condensed capsules into a collection of candidates $\left\{\hat \bu_{j|1}, \ldots, \hat \bu_{j|m} \right\}$ 
% used for representing the
%as to the 
for the 
$j$-th high-level capsule in layer $\ell+1$. 
Following~\citet{sabour2017dynamic}, each element $\hat \bu_{j|i}$ is calculated by: 
% , we use linear extrapolation by multiplying $\bu_i$ with a linear transformation matrix $\mW_j^c \in \mathbb R^{d\times d}$: 
% One simple solution to this is multiplying $\bu_i$ with a linear transformation matrix $\mW_j^c \in \mathbb R^{d\times d}$:
% by multiplying $u_i$ with a transformation matrix $W_j^c \in \mathbb R^{d\times d}$ as follows:
\vspace{-0.02in}
\begin{align*}
\hat \bu_{j|i}=\mW^c \bu_{i} \in \mathbb R^d  
\end{align*}
% \vspace{0.02in}
where $\mW^c$ is a linear transformation matrix.

Then, we represent a high-level capsule $\bv_j$ by %taking 
a weighted sum over those candidates,
% one capsule is calculated by using a weighted sum over those candidates and used for representing a high-level capsule $\bv_j$, 
denoted as:
\vspace{-0.19in}
% The basic idea of routing is to find 

% find multiple reliable capsules
% from a collection of capsule predictions, via clustering, and pass them into the subsequent layer
% construct a non-linear map in an iterative manner to output high-level capsules in the subsequent layer, via clustering: %\todo{SE: what is H?}
% \vspace{-0.2in}
% \begin{align*}
% % \left\{\hat \bu_{j|i}\in \mathbb R^d\right\}_{i=1,\ldots,m,j=1\ldots,n} \mapsto \left\{\bv_j\in \mathbb R^d\right\}_{j=1}^n.
% % \vspace{-0.3in}
% \bv_j =(\hat \bu_{j|1}, \dots, \hat \bu_{j|m})
% \end{align*}
\begin{align*}
\bv_j = \sum_{i=1}^m c_{ij} \hat \bu_{j|i}
\end{align*}
where $c_{ij}$ is a coupling coefficient iteratively updated by a clustering-like method.

\paragraph{Our Routing}

% we observe that the Capsule-B with 3 or 5 iterations of routing optimizes the loss faster and converges to a lower loss at the end than 1 iteration
As discussed earlier, routing algorithms like dynamic routing~\citep{sabour2017dynamic} and EM routing~\citep{hinton2018matrix}, 
%~\cite{sabour2017dynamic, zhao2018investigating, zhang2018fast} 
which use the same number of iterations for all samples, 
% with a fixed number of iterations 
perform well according to training loss at system level, %by taking 
% in a fixed number of iterations, 
but on instance level for individual examples, 
%we observe that desirable 
convergence has still not been reached. %which
This 
% individual examples have not converged.
% at instance-level have not converged. 
%raises 
increases 
the risk of 
unreliability 
% undesirable non-convergence 
for routing processes (see Figure~\ref{fig:evaluate-routing}).
To evaluate the performance of routing processes at instance level, we 
% provide an objective function by 
formulate them as a proxy problem minimizing 
% \todo{SE: needs an article in non-Chinese-English. The article must either be definite (the NAS function) or indefinite (a NAS function)} 
the negative agreement score (NAS) function:
% , which we call negative agreement score (NAS) function, such that the f 
% such that the following proxy function $\f$ (negative agreement score (NAS) function)\todo{SE: which is or which we call} 
% is minimized
\begin{align*}
&\min_{c, \bv} \: f(\bu) = - \sum_{i,j}c_{ij}\langle \bv_j, \bu_{j|i}\rangle\\ 
% \fk(\fd(\bv_j, \bu_{j|i}))\\
& \text{s.t.}\quad\forall{i,j}: c_{ij}>0, \quad \sum\limits_{j}c_{ij}=1.
% \label{eq:proxy-function}
\end{align*}

The basic intuition behind this is %one should always assigns 
to assign 
higher weights $c_{ij}$ to one agreeable pair $\langle \bv_j, \bu_{j|i}\rangle$ if the capsule $\bv_j$ and $\bu_{j|i}$ are close to each other 
% \todo{SE: agreeable? Wei: fixed} 
such that the total agreement score $\sum_{i,j}c_{ij} \langle \bv_j, \bu_{j|i}\rangle$ is maximized.
% Intuitively, 
% $f(v,c)$ reflects the average distance between high-level capsule $v_j$ and surrounding ``prediction capsules'' $\hat u_{j|i}$. A shorter distance means that high-level capsules are aggregated well by high-dimensional agreement. 
However, the choice of NAS functions remains an open problem.
\citet{hinton2018matrix} hypothesize that the agreeable pairs in NAS functions are from Gaussian distributions. Instead, we study NAS functions by introducing Kernel Density Estimation (KDE) since %it is a 
this yields a 
non-parametric density estimator requiring no
assumptions that the agreeable pairs are drawn from parametric distributions. Here, we formulate the NAS function in a KDE form.
% introduce a non-parametric clustering with Kernel Density Estimation (KDE) for since KDE techniques bridge a family of kernel functions and underlying empirical distributions, which often leads to computational efficiency, defined as:
% introduce an iterative routing process with kernel density estimation (KDE), where the KDE loss function is:
% Following~\citet{zhang2018fast},\todo{SE: reduce many self-references} we use the vanilla routing process \todo{SE: of Sabour et al.?} in relation with kernel density estimation (KDE) as an explicit loss function:
\begin{eqnarray}
%\max_{\mathbf v,\mathbf r} && 
%\max_{\mathbf v,\mathbf c} 
\min_{c, \bv} \: f(\bu) = - \sum_{i,j}c_{ij}k(d(\bv_j, \bu_{j|i}))
\label{eq:kde_loss}
\end{eqnarray}
where
% $z$ is a normalization constant, 
% $c_{ij}$ is the coupling coefficient that measures the intensity of connection between one ``prediction capsule'' $\hat u_{j|i}$ and aggregated high-level capsule $v_j$, 
% $\f(\bu)$ reflects the average distance between high-level capsule $\bv_j$ and surrounding \textit{prediction capsules} $\bu_{j|i}$, 
$d$ is a distance metric with $\ell_2$ norm, and $k$ is a Epanechnikov kernel function~\cite{wand1994kernel} with:
\vspace{-0.05in}
$$k(x)=
\begin{cases}
1 - x & x \in [0,1)\\
0 & x \geq 1
\end{cases}$$
% Intuitively, $\f(\bu)$ reflects the average distance between high-level capsule $\bv_j$ and surrounding ``prediction capsules'' $\bu_{j|i}$. 
% A shorter distance means that high-level capsules are aggregated well by high-dimensional agreement. 

% Intuitively, $f(v,c)$ shows the selection confidence of high-level capsules that stand out of a set of prediction capsules by strong agreement.\todo{SE: what is selection confidence?}\wei{Consider clustering, the confidence is like how well the centroids represent the members of their clusters. (residual sum of squares)}

% the average distance between high-level capsule $v_j$ and surrounding ``prediction capsules'' $\hat u_{j|i}$.
%To alternative optimization, 
The solution we used for KDE is taking Mean Shift~\cite{comaniciu2002mean} to minimize the NAS function $f(\bu)$:
\begin{eqnarray}
	\nabla f(\bu)= \sum_{i,j} c_{ij} k'(d(\bv_j, \bu_{j|i}))\frac{\partial{d(\bv_j, \bu_{j|i})}}{\partial{\bv}} \nonumber
\end{eqnarray}
First, $\bv_j^{\tau+1}$ can be updated while $c_{ij}^{\tau+1}$ is fixed:
\begin{align*}
	\bv_j^{\tau+1} = \frac{\sum_{i,j} c_{ij}^\tau k'(d(\bv_j^\tau, \hat \bu_{j|i})) \bu_{j|i}}{\sum_{i,j} k'(d(\bv_j^\tau, \bu_{j|i}))}
\end{align*}
Then, $c_{ij}^{\tau+1}$ can be updated using standard gradient descent: 
\begin{align*}
	c_{ij}^{\tau+1} = c_{ij}^\tau + \alpha \cdot k(d(\bv_j^\tau, \bu_{j|i}))
\end{align*}
where $\alpha$ is the hyper-parameter to control step size.

% Consider the convergence of individual examples not reaching,
To address the issue of convergence not being reached at instance level,
% To improve the convergence of routing processes at instance-level, 
we present an adaptive optimizer to self-adjust the number of iterations for individual examples according to their negative agreement scores
% depending on how routing processes perform at instance-level according to a negative agreement score 
(see Algorithm~\ref{alg:1}).
% a negative agreement score (see Algorithm~\ref{alg:1}).
% avoid selecting the number of iteration by observing the training loss at system-level, 
% by avoiding a fixed numbers of iteration used for routing process,
% fixed number of iterations used for routing process
% To adapt fixed number of iterations to various examples,
% we introduce adaptive optimizer by using %adaptive 
% variable times of iterations for each example.
% As shown in Figure~\ref{fig:kde_curve}, convergence rate is different for various examples, so the adaptive times of iterations %shall be considered.
% is intuitively beneficial. 
Following~\citet{zhao2018investigating}, we replace standard softmax with leaky-softmax, which decreases the strength of noisy capsules. 
% Our complete routing process is summarized in Algorithm~\ref{alg:1}. 

% In addition, we replace standard softmax with Leaky-Softmax~\cite{zhao2018investigating}, which decreases the strength of noisy capsules.
% reduces the influence of noise surrounding capsules. 
% to aggregated high-level capsules.

% while updating the intensity of connection $c_{ij}$ between aggregated high-level capsule and surrounding ``prediction capsules'', which reduces the influence of noise surrounding capsules to the winning capsules.
% The detailed derivation is provided in the appendix. 

\begin{algorithm}
\caption{Our Adaptive KDE Routing}\label{alg:1}
\begin{algorithmic}[1]
% \small
\State \textbf{procedure} ROUTING($\bu_{j|i}$, $\ell$)
\State Initialize $\forall i,j: c_{ij}=1/n_{\ell+1}$
\While {true}
 \ForAll{capsule $i$, $j$ in layer $\ell$, $\ell+1$}
  \State $c_{ij} \gets \text{leaky-softmax}(c_{ij})$
 \EndFor  
 
 \ForAll{capsule $j$ in layer $\ell+1$}
  \State $\bv_j \gets \frac{\sum_{i} c_{ij} k'(d(\bv_j, \bu_{j|i})) \hat \bu_{j|i}}{\sum_{i=1}^n k'(d(\bv_i, \bu_{j|i}))}$
 \EndFor  

 \ForAll{capsule $i$, $j$ in layer $\ell$, $\ell+1$}
  \State $c_{ij}\gets c_{ij} + \alpha \cdot k(d(\bv_j, \bu_{j|i}))$
 \EndFor 
 
 \ForAll{capsule $j$ in layer $\ell+1$}
  \State $a_j\gets |v_j|$ 
 \EndFor
 
 \State $\rm{NAS} = \log(\sum_{i,j}c_{ij}k(d(\bv_j, \bu_{j|i})))$
 \If{$|\rm{NAS}- \rm{Last\_NAS}| < \epsilon$}
   \State \textbf{break}
 \Else
   \State $\rm{Last\_NAS} \gets \rm{NAS}$
 \EndIf
\EndWhile
\State \textbf{return} $v_j$, $a_j$
\end{algorithmic}
\end{algorithm}

\subsection{Representation Layer}
This is the top-level layer containing final capsules calculated by iteratively minimizing the NAS function (See Eq.~\ref{eq:kde_loss}), where the number of final capsules corresponds to the entire output space. Therefore,
% This layer contains final capsules calculated by iteratively minimizing NAS function (See Eq. ).
% input sentences or documents are encoded into final capsules,
% representations 
% i.e., label capsules in text classification task.
% , or question and answer capsules in QA task. 
% The number of final capsules corresponds to the size of an output space, i.e., the number of labels in text classification. Each final capsule is the centroid of cluster 
% each final capsule stands for a high-level capsule in the top-level layer, calculated by iteratively minimizing NAS function (See Eq. ).
as long as the size of an output space goes to a large scale (thousands of labels), the computation of this function would become extremely expensive, which yields the bottleneck of scalability of capsule networks.

% leading to the bottleneck of scalability of capsule networks, the size of an output space goes to a large scale (thousands of labels). Such issue yields the bottleneck of the scalability of capsule networks.

% which yields the bottleneck of scalability of capsule networks as long as the size of an output space goes to a large scale (thousands of labels). 
% comes to be a large scale (thousands of labels). 

\paragraph{Partial Routing} As opposed to the entire output space on data sets, the sub-output space corresponding to individual examples is rather small, i.e., only few labels are assigned to one document in text classification, for example. As a consequence, it %comes to be 
is 
redundant to route low-level capsules to the entire output space for each example 
% as we know the labels assignment 
in the training stage, which motivated us to
% Such case makes it redundant to route low-level capsules as to the entire output space for each example. we 
present a partial routing algorithm with constrained output spaces, such that our NAS function is described as:
\begin{align*}
\min_{c, \bv} \: -\sum_i(\sum_{j\in {D^{+}}}c_{ij}\langle \bv_j, \bu_{j|i}\rangle\\
+\lambda \cdot \sum_{k \in {D^{-}}}
c_{ik}\langle \bv_k, \bu_{k|i}\rangle)
% \min_{c, \bv} \: -\sum_i\sum_{j\in {D^{+}\cup D^{-}}}c_{ij}\langle \bv_j, \bu_{j|i}\rangle
% - \sum_i(\sum_j c_{ij}\fk(\fd(\bv_j, \bu_{j|i})) + \sum_j c_{ij}\fk(\fd(\bv_j, \bu_{j|i})))
\end{align*}
where $D^{+}$ and $D^{-}$ denote the sets of real (positive) and randomly selected (negative) outputs for each example, respectively. %The scale of 
%Combining 
Both %two 
sets %comes to be far 
are far smaller than the entire output space.

\section{Experiments}\label{sec:setup}
The major focus of this work is to investigate the scalability of our approach on datasets with a large output space, and generalizability in low-resource settings with few training examples. 
%training examples condition.
% data settings. 
Therefore, we validate our capsule-based approach on two specific NLP tasks: (\romannumeral1) multi-label text classification with a large label scale; (\romannumeral2) question answering %with heavy label imbalanced issue, 
with a data imbalance issue.
% since
% We choose these because 
%ince
% By two tasks,
% our major focus is to study the scalability of our model on datasets with large output space, and generalization capability in small 
% %training examples condition.
% data settings. 

%Each application has its own background, experiments and discussions.

\subsection{Multi-label Text Classification}
Multi-label text classification task %which 
refers to assigning %the 
multiple 
relevant labels to each input document, while the entire label set might be extremely large. %is important in NLP tasks.
%Formally, 
% Given an input document, we use our capsule framework to generate final capsules as to labels in the representation layer. 
We use our approach to encode an input document and generate the final capsules corresponding to the number of labels in the representation layer. 
% $v_j \in \mathbb{R}^d$ for each label $j$ in the representation layer.
% \todo{SE: I assume you have nodes or capsules in the representation layer, but not labels ... Wei:fixed} 
The length of final capsule %as to 
for each label indicates the probability 
% of the case 
whether the document %belongs to 
has this label.

\begin{table}[htbp]
\centering
	\resizebox{0.8\columnwidth}{!}{
% 	{\foonotesize
	\begin{tabular}{|l|r r |}
		\toprule
		Dataset & \#Train/Test/Labels & Avg-docs\\
		\midrule
    RCV1 & 23.1K/781.2K/103 & 729.67\\
		EUR-Lex & 15.4K/3.8K/3.9K & 15.59\\
		\bottomrule
	\end{tabular}}
	\caption{Characteristics of the datasets. Each label of RCV1 has about 729.67 training examples, while each label of EUR-Lex has merely about 15.59 examples.}

	\label{tab:data}
	\vspace{-0.1in}
\end{table}

\paragraph{Experimental Setup}
We conduct our experiments 
on 
% text classification benchmarks by selecting 
two datasets selected from the extreme classification repository:\footnote{https://manikvarma.github.io \label{repo}} a regular label scale dataset (RCV1), with 103 labels~\cite{lewis2004rcv1}, and a large label scale dataset (EUR-Lex), with 3,956 labels~\cite{mencia2008efficient}, %illustrated 
described 
in Table~\ref{tab:data}. 
The intuition behind our datasets selection is that EUR-Lex, with 3,956 labels and 15.59 examples per label, %is accord 
fits well 
with our goal of investigating the scalability and generalizability of our approach. We contrast EUR-Lex %dataset 
with RCV1, a dataset with a regular label scale, and 
% evaluating the generalization capability of capsule networks in low-resource setting with few training examples.
% We select EUR-Lex rather than other multi-label datasets such as Amazon-12K with 12,277 labels, since one of our major foci is to investigate how our capsule model performs in small %training examples
% data 
% conditions %like EUR-Lex 
% with few examples (15.59 in EUR-Lex) per label.
% EUR-Lex has fewer examples (15.59) per label and investigating how our capsule model performs in few training examples condition is one of our major focus. 
leave the study of datasets with extremely large labels, e.g., Wikipedia-500K with 501,069 labels, to future work.

\paragraph{Baselines} We compare our approach to the following baselines: %including 
non-deep learning approaches using TF-IDF features of documents as inputs: FastXML~\cite{prabhu2014fastxml}, %SLEEC~\cite{bhatia2015sparse} 
and PD-Sparse~\cite{yen2016pd},
% to represent input documents, 
deep learning approaches using raw text of documents as inputs: FastText~\cite{joulin2016bag}, Bow-CNN~\cite{johnson2014effective}, CNN-Kim~\cite{kim2014convolutional}, XML-CNN~\cite{liu2017deep}), and a capsule-based approach Cap-Zhao~\cite{zhao2018investigating}. 
% The raw text of documents on these datasets and TF-IDF features can be found in the Extreme Classification Repository. 
For %the evaluation of those compared methods, 
evaluation, 
we use standard rank-based measures~\citep{liu2017deep} such as Precision@k, and Normalized Discounted Cumulative Gain (NDCG@$k$). %\textsuperscript{\ref{repo}}.

\paragraph{Implementation Details} The word embeddings are initialized as 300-dimensional GloVe vectors~\cite{pennington2014glove}. In the convolutional layer, we use a convolution operation with three different window sizes 
% of the convolution kernel is set to 
(2,4,8) to extract features from input documents. Each feature is transformed into a primary capsule with 16 dimensions by a group-convolution operation. All capsules in the primary capsule layer are condensed into 256 capsules for RCV1 and 128 capsules for EUR-Lex %dataset 
by a capsule compression operation.

To avoid routing low-level capsules to the entire label space in the inference stage, we use a CNN baseline~\citep{kim2014convolutional} trained on the same dataset with our approach, to generate 200 candidate labels and take these labels as a constrained output space for each example.

%RESTORE-BEGIN
\begin{table*}[htbp]
	\centering
	\resizebox{2.1\columnwidth}{!}{
	\begin{tabular}{|l|l|c|c|c|c|c|c|c|c|r|}
		\toprule
		 \textbf{Datasets} & \textbf{Metrics}& \textbf{FastXML} & 
		 \textbf{PD-Sparse} & \textbf{FastText}& \textbf{Bow-CNN} & \textbf{CNN-Kim} & \textbf{XML-CNN} & \textbf{Cap-Zhao}& \textbf{NLP-Cap} & \textbf{Impv} \\
		\midrule
		\multirow{3}{*}{RCV1}& 
		 PREC@1 & 94.62 
		 & 95.16 & 95.40 & 96.40 & 93.54 &96.86 & 96.63 & \textbf{97.05} &+0.20\% \\ 
		 & PREC@3 & 78.40 
		 & 79.46 & 79.96 & 81.17 & 76.15 & 81.11 & 81.02 & \textbf{81.27} &+0.20\% \\ 
		 & PREC@5 & 54.82 
		 & 55.61 & 55.64 & \textbf{56.74} & 52.94 & 56.07 & 56.12 & 56.33 &-0.72\%\\ 
		 & NDCG@1 & 94.62 
		 & 95.16 & 95.40 & 96.40 & 93.54 & 96.88 & 96.63 & \textbf{97.05} &+0.20\%\\ 
		 & NDCG@3 & 89.21 
		 & 90.29 & 90.95 & 92.04 & 87.26 & 92.22 & 92.31 & \textbf{92.47} &+0.17\%\\ 
		 & NDCG@5 & 90.27 
		 & 91.29 & 91.68 & 92.89 & 88.20 & 92.63 & 92.75 & \textbf{93.11} &+0.52\%\\ 
		\midrule
		
    \multirow{3}{*}{EUR-Lex}& 
		  PREC@1 & 68.12 
		  & 72.10 & 71.51 & 64.99 & 68.35 & 75.65 & - & \textbf{80.20} & +6.01\% \\
		 & PREC@3 & 57.93 
		 & 57.74 & 60.37 & 51.68 & 54.45 & 61.81 & - & \textbf{65.48} & +5.93\% \\
		 & PREC@5 & 48.97 
		 & 47.48 & 50.41 & 42.32 & 44.07 & 50.90 & - & \textbf{52.83} & +3.79\%\\
		 & NDCG@1 & 68.12 
		 & 72.10 & 71.51 & 64.99 & 68.35 & 75.65 & - & \textbf{80.20} & +6.01\% \\
		 & NDCG@3 & 60.66 
		 & 61.33 & 63.32 & 55.03 & 59.81 & 66.71 & - & \textbf{71.11} & +6.59\% \\
		 & NDCG@5 & 56.42 
		 & 55.93 & 58.56 & 49.92 & 57.99 & 64.45 & - & \textbf{68.80} & +6.75\% \\
		\bottomrule
	\end{tabular}
	}
  \caption{Comparisons of our NLP-Cap approach and baselines on two text classiï¬cation benchmarks, where '-' denotes methods that failed to scale due to memory issues.}
  \label{tab:experiment}
  \vspace{-0.05in}
\end{table*}
%RESTORE-END

\paragraph{Experimental Results}
In Table~\ref{tab:experiment},
we can see a noticeable margin brought by our capsule-based approach over the strong baselines on EUR-Lex, and competitive results on RCV1. 
%We conjecture that our approach 
These results appear to indicate that our approach 
has superior generalization ability 
on datasets with fewer training examples, i.e., RCV1 has 729.67 examples per label while EUR-Lex has 15.59 examples.

In contrast to the strongest baseline XML-CNN with 22.52M parameters and 0.08 seconds per batch, our approach has 14.06M parameters, and takes 0.25 seconds in an acceleration setting with capsule compression and partial routing, and 1.7 seconds without acceleration.
% Regarding number of parameters, the strong baseline XML-CNN has 22.52M whereas our approach has 14.06M parameters. In terms of computational efficiency, XML-CNN takes 0.08 seconds for one batch, while our NLP-Capsule approach takes 0.25 seconds with fast training strategies and 1.7 seconds without.\todo{SE: with the fast training strategy?} % Therefore, we 
This demonstrates that our approach 
% needs fewer parameters and 
provides competitive computational speed with fewer parameters compared to the competitors.

\begin{figure}[t]
\centering
\subfigure{\includegraphics[width=0.49\columnwidth]{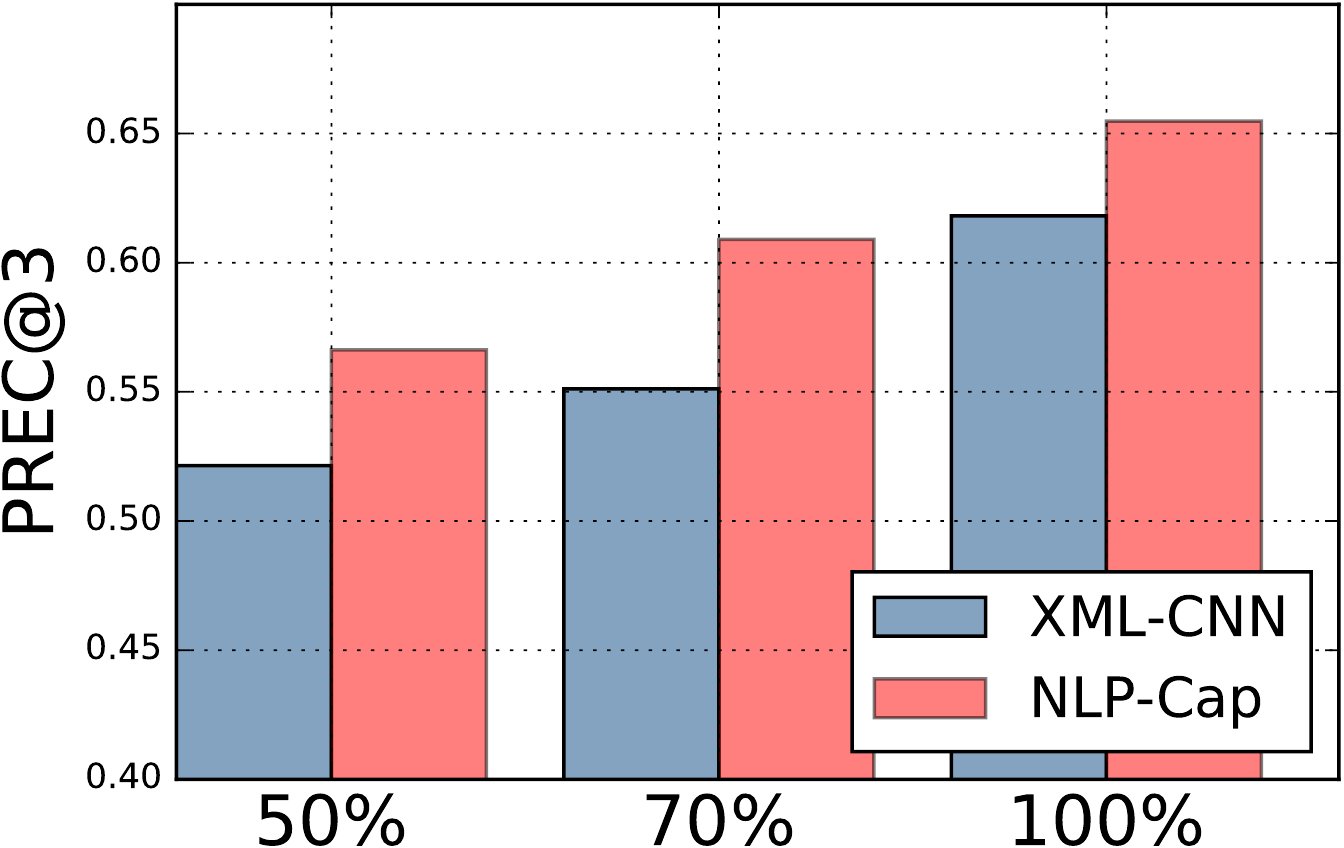}}
\subfigure{\includegraphics[width=0.49\columnwidth]{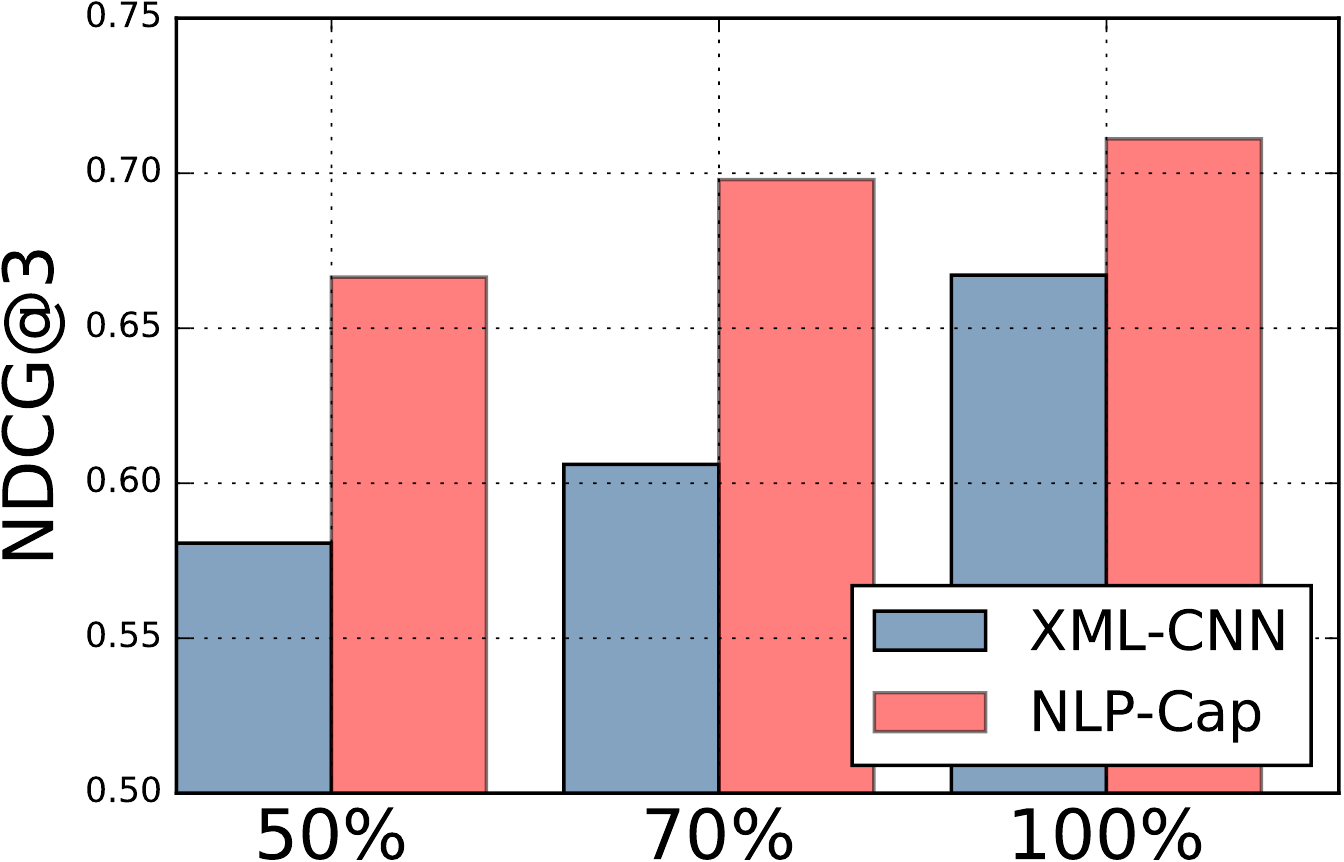}}
\caption{Performance on EUR-Lex by varying the percentage of training examples (X-axis).}\label{fig:percentage}
\vspace{-5pt}
\end{figure}

\paragraph{Discussion on Generalization}
To further study the generalization capability of our approach, we 
% investigate our model's capability to deal with such scenarios by 
vary the percentage of training examples from 100\% to 50\% on the entire training set, 
%while the 
leading to 
% an 
the 
% average 
number of training examples per label 
decreasing
%is 
from 15.59 to 7.77. Figure~\ref{fig:percentage} shows that our approach %achieves significant improvement 
%sometimes substantially 
outperforms 
%on 
the strongest baseline XML-CNN with different fractions of the training examples. 

This finding agrees with our speculation on generalization:
% Congruent with our reasoning about generalization, 
the distance between our approach and XML-CNN increases as fewer training data samples are available. In Table~\ref{tab:percentage}, we also find that our approach with 70\% of training examples achieves about 5\% improvement over XML-CNN with 100\% of examples on 4 out of 6 metrics.

%RESTORE-BEGIN

\begin{table}[t]
\centering
\resizebox{1\columnwidth}{!}{
\begin{tabular}{|l|c|c|c|c|}
\toprule
Method & \#Training & PREC@1 & PREC@3 & PREC@5 \\
\midrule
\multirow{1}{*}{XML-CNN}
&100\% examples & 75.65 & 61.81 & 50.90 \\ 
\midrule
\multirow{3}{*}{NLP-Capsule}&
50\% examples & 73.69 & 56.62 & 44.36 \\
&60\% examples & 74.83 & 58.48 & 46.33 \\
&70\% examples & 77.26 & 60.90 & 47.73\\ 
&80\% examples & 77.68 & 61.06 & 48.28 \\
&90\% examples & 79.45 & 63.95 & 50.90 \\
&100\% examples & \textbf{80.20} & \textbf{65.48} & \textbf{52.83}\\
\bottomrule
Method & \#Training & NDCG@1 & NDCG@3 & NDCG@5 \\
\midrule
\multirow{1}{*}{XML-CNN}
&100\% examples & 75.65 & 66.71 & 64.45 \\ 
\midrule
\multirow{3}{*}{NLP-Capsule}&
50\% examples & 73.69 & 66.65 & 67.36 \\
&60\% examples & 74.83 & 67.87 & 68.62 \\
&70\% examples & 77.26 & 69.79 & 69.65 \\ 
&80\% examples & 77.67 & 69.43 & 69.27 \\
&90\% examples & 79.45 & \textbf{71.64} & \textbf{71.06} \\
&100\% examples & \textbf{80.21} & 71.11 & 68.80 \\
\bottomrule
\end{tabular}
}
\caption{Experimental results on different fractions of training examples from 50\% to 100\% on EUR-Lex. \label{tab:percentage}}
\end{table}
%RESTORE-END

\paragraph{Routing Comparison}
We compare our routing with \cite{sabour2017dynamic} and \cite{zhang2018fast} on EUR-Lex dataset
% , as shown in Figure ~\ref{fig:routing-bar}, 
% To compare various routing algorithms, we provide overall performance of Sabour's~\cite{sabour2017dynamic}, Zhang's~\cite{zhang2018fast} and our routing on EUR-Lex dataset in Figure ~\ref{fig:routing-bar}. 
and observe that it performs best on all metrics (Table~\ref{tab:ablation}). We speculate that the improvement comes from enhanced reliability of routing processes at instance level.

\begin{table}[b]
	\centering
	\resizebox{1\columnwidth}{!}{
	\begin{tabular}{|l |c | c | c |}
		\toprule
     \textbf{Method}& \textbf{PREC@1} & \textbf{PREC@3} & \textbf{PREC@5} \\ 
     \midrule
    XML-CNN& 75.65 & 61.81 & 50.90 \\ 
    
    NLP-Capsule + Sabour`s Routing & 79.14 & 64.33 & 51.85 \\
    
    NLP-Capsule + Zhang`s Routing& 80.20 & 65.48 &52.83 \\
		\bottomrule
		NLP-Capsule + Our Routing & \textbf{80.62} & \textbf{65.61} &
    \textbf{53.66}\\

  	\toprule
     \textbf{Method}& \textbf{NDCG@1} & \textbf{NDCG@3} & \textbf{NDCG@5} \\ 
     \midrule
    XML-CNN& 75.65 & 66.71 & 64.45 \\ 
    
    NLP-Capsule + Sabour`s Routing & 79.14 & 70.13 & 67.02 \\
    
    NLP-Capsule + Zhang`s Routing& 80.20 & 71.11 & 68.80\\
		\bottomrule
		NLP-Capsule + Our Routing & \textbf{80.62} & \textbf{71.34} & \textbf{69.57}\\
    
  	\bottomrule
	\end{tabular}
	}
  \caption{Performance on EUR-Lex dataset with different routing process.}\label{tab:ablation}
  \vspace{-0.3cm}
\end{table}

\subsection{Question Answering}
Question-Answering (QA) selection task refers to selecting
% is a traditional ranking task, which aims to select 
the best answer from candidates to each question.
% aims to rank candidate answers that match a given question. 
%Let 
For a question-answer pair $(q,a)$, we use our capsule-based approach to generate two final capsules $\bv_q$ and $\bv_a$ corresponding to the respective question and answer.
% in the representation layer.
% yield question capsule $v_q \in \mathbb{R}^d$ and answer capsule representation $v_a \in \mathbb{R}^d$ in the representation layer,
%from one question $q$ and answer $a$, 
% where question $q$ and answer $a$ consist of sequences of words as inputs. 
The relevance score of question-answer pair 
% for ranking answer candidates 
can be defined as their cosine similarity:
% In question answering (QA) tasks, a question $q$ and answer $a$ are consisted of a sequence of words. QA tasks aim to understand natural language questions and then select the candidate answers that best match those questions. 
\begin{align*}
s(q,a) = \rm{cos}(\bv_q, \bv_a) = \frac{\bv_q^{\intercal} \bv_a}{||\bv_q|| \cdot ||\bv_a||}
\end{align*}
% The goal is a binary classification of a $(q,a)$ pair as positive or negative. 
\paragraph{Experiment Setup}
In Table~\ref{tab:qa_data}, we conduct our experiments on the TREC QA dataset collected from TREC QA track 8-13 data~\cite{wang2007jeopardy}. %since it 
The intuition behind this dataset selection is that the cost of hiring human annotators to collect positive answers for individual questions can be prohibitive since positive answers can be conveyed in multiple different surface forms. Such issue arises particularly in TREC QA with only 12\% positive answers. Therefore, we use this dataset to investigate the generalizability of our approach.

\paragraph{Baselines}We compare our approach to the following baselines: CNN + LR~\cite{yu2014deep} using unigrams and bigrams, CNN~\cite{severyn2015learning} using additional bilinear similarity features, CNTN~\cite{qiu2015convolutional} using neural tensor network, LSTM~\cite{tay2017learning} using single and multi-layer, MV-LSTM~\cite{wan2016deep}, NTN-LSTM and HD-LSTM~\cite{tay2017learning} using holographic dual LSTM and Capsule-Zhao~\cite{zhao2018investigating} using capsule networks.
% \todo{SE: how different to the current work? Wei: discussed in related work} 
For %the 
evaluation, 
%of those compared methods, 
we use standard measures~\cite{wang2007jeopardy} such as Mean Average Precision (MAP) and Mean Reciprocal Rank (MRR).

\begin{table}[t]
\centering
	\resizebox{1\columnwidth}{!}{
	\begin{tabular}{| c| c | c| c|}
		\toprule
		Dataset & \#Questions & \#QA Pairs & \%Positive \\
		\midrule
		Train/Dev/Test & 1229/82/100 & 53417/1148/1517 & 12\%\\
		\bottomrule
	\end{tabular}}
	\caption{Characteristic of TREC QA dataset. \%Positive denotes the percentage of positive answers.}
	\label{tab:qa_data}
\end{table}

\begin{table}[b]
\centering
\small
	\begin{tabular}{|l| c |c|}
		\toprule
		Method & MAP & MRR\\
		\midrule
		CNN + LR (unigram) & 54.70 & 63.29 \\
		CNN + LR (bigram) & 56.93 & 66.13 \\
		CNN & 66.91 & 68.80 \\
		CNTN & 65.80 & 69.78 \\
		\midrule
		LSTM (1 layer) & 62.04 & 66.85 \\
		LSTM & 59.75 & 65.33 \\
		MV-LSTM & 64.88 & 68.24 \\
		NTN-LSTM & 63.40 & 67.72 \\
		HD-LSTM & 67.44 & \textbf{75.11} \\
		\midrule
		Capsule-Zhao & 73.63 & 70.12 \\
		\midrule
		NLP-Capsule & \textbf{77.73} & 74.16\\       
		\bottomrule
	\end{tabular}
	\caption{Experimental results on TREC QA dataset.}
	\label{tab:qa_results}
\end{table}
%RESTORE-END

\paragraph{Implementation Details} The word embeddings used for question answering pairs are initialized as 300-dimensional GloVe vectors. In the convolutional layer, we use a convolution operation with three different window sizes (3,4,5). All 16-dimensional capsules in the primary capsule layer are condensed into 256 capsules by the capsule compression operation.

\paragraph{Experimental Results and Discussions}

In Table~\ref{tab:qa_results}, 
the best performance on MAP metric is achieved by our approach,

which verifies the effectiveness of our model. 
We also observe that our approach exceeds traditional neural models like CNN, LSTM and NTN-LSTM by a noticeable margin. 

This finding also agrees with the observation we found in multi-label classification: our approach has superior generalization capability in low-resource setting with few training examples.

In contrast to the strongest baseline HD-LSTM with 34.51M and 0.03 seconds for one batch, our approach has 17.84M parameters and takes 0.06 seconds in an acceleration setting, and 0.12 seconds without acceleration.

\section{Related Work}\label{sec:related}
%In this section, we review recent progress on multi-label text classification, question answering, and capsule networks.

\subsection{Multi-label Text Classification}
 Multi-label text classification aims at assigning a document to a subset of labels whose label set might %contain a large number, 
 be extremely large. %, like from hundreds to thousands. 
 With increasing numbers of labels, %they are facing challenges on 
 issues of data sparsity and scalability arise. 
 %As seen in many recent works, various tree-based, embedding-based and deep learning models are devised.
 Several methods have been proposed for the large multi-label classification case. 
 
%The first type, 
\textbf{Tree-based models}~\cite{agrawal2013multi,weston2013label} induce a tree structure that recursively partitions the feature space with non-leaf nodes. %Meanwhile, 
Then, the restricted label spaces at leaf nodes are used for classification.
% At each leaf node, a classifier only classifies within a %limited 
% restricted label space. 
Such a solution %brings advantages for higher
% sub-linear reduction of time complexity. 
entails higher robustness because of a dynamic hyper-plane design and its
computational efficiency. 
FastXML~\cite{prabhu2014fastxml} is one such tree-based model, %of 
% such model type 
which learns a hierarchy of training instances and optimizes an NDCG-based objective function for nodes in the tree structure.
% each node of the hierarchy. 
%The second type, 

\textbf{Label embedding models}~\cite{balasubramanian2012landmark,chen2012feature,cisse2013robust,bi2013efficient,ferng2011multi,hsu2009multi,ji2008extracting,kapoor2012multilabel,lewis2004rcv1,yu2014large} address the data sparsity issue with two steps: compression and decompression. The compression step learns a low-dimensional label embedding that is projected from original and high-dimensional label space. When data instances are classified to these label embeddings, they will be projected back to the high-dimensional label space, 
% in the target space, 
which is the decompression step. Recent works came up with different compression or decompression techniques, e.g., SLEEC~\cite{bhatia2015sparse}. %Among these, SLEEC~\cite{bhatia2015sparse} is one of the state of the art. 
% A third model class 
%embraces the techniques in recent development of natural language processing, namely 
% relies on 

\textbf{Deep learning models}: FastText~\cite{joulin2016bag} uses %classifies %the
averaged word embeddings %as 
to classify documents,
% document embeddings by taking the average over individual word embeddings.
% averaged word embeddings from documents, 
which is computationally efficient but 
%does not adequately leverage the context information like word ordering.
ignores word order. 
% take context information into account.\todo{SE: ? I would say it uses context information but not very well, as it averages everything}
Various CNNs inspired by~\citet{kim2014convolutional} explored MTC with dynamic pooling, such as Bow-CNN~\cite{johnson2014effective} and XML-CNN~\cite{liu2017deep}. 

\textbf{Linear classifiers}: PD-Sparse~\cite{yen2016pd} introduces a Fully-Corrective Block-Coordinate Frank-Wolfe algorithm to address %issues of 
data sparsity. 

% Finally, there are also \textbf{linear classifiers} for each label, e.g., PD-Sparse~\cite{yen2016pd}. %belongs to learns a linear classifier for each label. 
% To address the sparsity issue, it introduces a Fully-Corrective Block-Coordinate Frank-Wolfe training algorithm.

%Unfortunately, few of the above works address the issue on insufficient number of examples for tail labels.

%the imbalance issue of data instances of different labels.
\subsection{Question and Answering}
%Question and answering (QA) is a task to find the corresponding answer to a factoid question. 
State-of-the-art approaches to QA fall into two categories: IR-based and knowledge-based QA.

\textbf{IR-based QA} firstly preprocesses the question and employ information retrieval techniques to retrieve a list of %most 
relevant passages to questions. Next, reading comprehension techniques are adopted to extract answers within the span of retrieved text. For answer extraction, early methods manually designed patterns to get them~\cite{pacsca2003open}. A recent popular trend is neural answer extraction. Various neural network models are employed to represent questions~\cite{severyn2015learning,wang2015long}. Since the attention mechanism naturally explores relevancy, it has been widely used in QA models to relate the question to candidate answers 
% in the text 
\cite{tan2016improved,santos2016attentive,sha2018multi}. Moreover, some researchers leveraged external large-scale knowledge bases to assist answer selection~\cite{savenkov2017evinets,shen2018knowledge,deng2018knowledge}. 

\textbf{Knowledge-based QA} conducts semantic parsing on questions and transforms parsing results into logical forms.
% , which is usually a logical tuple. 
Those forms are adopted to match answers from structured knowledge bases 
% Candidate answers are ranked by descending similarity score, and then highest answers selected 
\cite{yao2014information,yih2015semantic,bordes2015large,yin2016simple,hao2017end}. Recent developments focused on modeling 
the interaction between question and answer pairs: Tensor layers~\cite{qiu2015convolutional,wan2016deep} and holographic composition~\cite{tay2017learning} have pushed %the bar to 
the state-of-the-art.

%\textbf{Hybrid methods} integrate both the IR-based methods with knowledge-based together to generate answers, such as the DeepQA system\cite{allen2019enhanced} from IBM Watson.

\subsection{Capsule Networks}
Capsule networks were initially proposed by Hinton~\cite{hinton2011transforming} to improve representations learned by neural networks against %the 
vanilla CNNs. %and it also helps enhance the robust of network. 
Subsequently, \citet{sabour2017dynamic} replaced the scalar-output feature detectors of CNNs with vector-output capsules and max-pooling with routing-by-agreement.

\citet{hinton2018matrix} then proposed a new iterative routing procedure between capsule layers based on the EM algorithm, which achieves better accuracy on the smallNORB dataset. 
\citet{zhang2018attention} applied capsule networks to relation extraction in a multi-instance multi-label learning framework.~\citet{xiao2018mcapsnet} explored capsule networks for multi-task learning.

\citet{xia2018zero} studied the zero-shot intent detection problem with capsule networks, which aims to detect emerging user intents in an unsupervised manner.
\citet{zhao2018investigating} investigated capsule networks with dynamic routing for text classification, and transferred knowledge from the single-label to multi-label cases.~\citet{Cho:2019} studied capsule networks with determinantal point processes for extractive multi-document summarization.

Our work is different from our predecessors in the following aspects: (\romannumeral1) we evaluate the performance of routing processes at instance level, and introduce an adaptive optimizer to enhance the reliability of routing processes; 
(\romannumeral2) we present capsule compression and partial routing to achieve better scalability of capsule networks on datasets with a large output space.

\section{Conclusion}\label{sec:conclusion}
% Making computers perform in a more human-like manner is a major issue in current artificial intelligence (AI) and NLP research. This not only includes making computers perform on similar levels~\cite{Hassan:2018}, but also requesting them 
% to be robust to adversarial examples~\cite{Eger:2019} and generalizing from few data points~\cite{Rueckle:2019}. 
Making computers perform more like humans is a major issue in NLP and machine learning. This not only includes making them perform on similar levels \cite{Hassan:2018}, but also requests them 
to be robust to adversarial examples \cite{Eger:2019} and generalize from few data points \cite{Rueckle:2019}. In this work, we have addressed the latter issue.  

In particular, we
%provide an alternative account of thinking in generalization capability with extrapolations and high-dimensional agreement, 
%and 
extended existing capsule networks 
% into a new framework 
% being more suitable 
% for NLP tasks.
% with advantages concerning concerning scalability, reliability and generalizability, which is of being suitable for NLP tasks now
% a unified NLP-Capsule framework 
% with advantages concerning scalability, reliability and generalizability.
% stability, scalability and generalization. 
%In the end, we 
% Our capsule-based framework, 
into a new framework with advantages concerning scalability, reliability and generalizability. 
Our experimental results have demonstrated its effectiveness on two NLP tasks: %such as
multi-label text classification and question answering.
% , especailly the superior generalization capability in low-resource setting with few training examples.
% Applying our capsule-based framework on two NLP tasks, %such as
% large multi-label text classification and question answering, shows its effectiveness over strong baselines. 

Through our modifications and enhancements, we hope to have made capsule networks more suitable to large-scale problems and, hence, more mature for real-world applications.
In the future, %, further experiments on more tasks 
we plan to apply capsule networks to even more challenging NLP problems such as language modeling and text generation.

% sentiment analysis, and dialogue systems. 

\section{Acknowledgments}
We thank the anonymous reviewers for their comments, which greatly improved the final version of the paper. 
This work has been supported by the German Research Foundation as part of the Research Training
Group Adaptive Preparation of Information from Heterogeneous Sources (AIPHES) at the Technische
Universit\"at Darmstadt under grant No. GRK 1994/1.

\bibliography{acl2019}
\bibliographystyle{acl_natbib}

\clearpage

\end{document}